\documentclass[11pt]{article}

\PassOptionsToPackage{hyperfootnotes=false}{hyperref}
\usepackage[preprint]{acl}

\usepackage{times}
\usepackage{latexsym}

\usepackage[T1]{fontenc}

\usepackage[utf8]{inputenc}

\usepackage{microtype}

\usepackage{inconsolata}

\usepackage{graphicx}

\usepackage{booktabs}

\title{DiffResearch: Accelerating Deep Research with Diffusion Language Models\thanks{Accepted to EMNLP 2026 Industry Track.}}

\author{
  Alan-Barsag Gazzaev$^{1}$ \quad
  Alexey Gavrilov$^{1}$ \quad
  Mikhail Mozikov$^{2}$ \quad
  Sergey Muravyov$^{1}$ \\
  $^{1}$ITMO University, Kronverksky Pr. 49, bldg. A, St. Petersburg, 197101, Russia \\
  $^{2}$AXXX, 5 Nizhny Susalny Pereulok, bldg. 19, Moscow, 105064, Russia \\
  \texttt{avgazzaev@itmo.ru, aogavrilov@itmo.ru, mozikov@axxx.tech, smuravyov@itmo.ru}
}

\begin{document}
\maketitle
\begin{abstract}
Deep research systems iteratively plan, gather evidence, synthesize, and revise, making end-to-end latency a key bottleneck for interactive use. Almost all production systems are built on autoregressive (AR) language models and inherit left-to-right decoding in planning and long-form writing. We present DiffResearch, an open multi-agent deep research framework that places a diffusion language model (dLLM) at the core of every stage, and measure how a diffusion backbone changes latency and report quality. Holding the agentic scaffold, retrieval code, prompts, and evaluation fixed, we swap the backbone between an open diffusion model and a comparable AR model. Across 100 Deep Research Bench tasks, the diffusion backbone is roughly $6\times$ faster end-to-end, driven by order-of-magnitude reductions in planning and writing. We further introduce dFast, a decomposition mode that reads intermediate denoising states and extracts subqueries before generation finishes, providing a $1.99\times$ median speedup for decomposition in isolation with no loss of report quality. On the benchmark's RACE rubric, dFast obtains the best overall score among the three configurations. Finally, in a public system-level leaderboard snapshot, an instance built on a stronger diffusion model ranks above several commercial systems. We release DiffResearch under an Apache-2.0 license.
\end{abstract}

\section{Introduction}

The past two years have seen deep research (DR) systems emerge as a flagship application of large language models (LLMs). Commercial products such as OpenAI Deep Research \citep{openai2025deepresearch}, Gemini Deep Research \citep{google2024gemini}, and Perplexity Deep Research \citep{perplexity2025deepresearch}, alongside a growing body of open-source frameworks \citep{langchain2025odr,zheng2025deepresearcher}, autonomously orchestrate multi-step web exploration, targeted retrieval, and higher-order synthesis to turn large amounts of online information into citation-rich reports. Among the tasks these systems perform, scientific literature review is one of the most demanding: a review requires balanced thematic coverage, faithful attribution, the identification of competing methodological camps, and a coherent narrative that situates each work within the broader landscape. As Deep Research Bench \citep{li2025deepresearchbench} and similar evaluations \citep{du2025deepresearchgym,zheng2025researchqa} have shown, the strongest agents now approach the quality of human analysts on PhD-level tasks across dozens of fields.

Despite this progress, almost all production DR agents share one architectural backbone: they are built on autoregressive LLMs \citep{brown2020language,touvron2023llama,openai2023gpt4} and inherit the same left-to-right, single-pass decoding that governs their token generators. This shapes both \emph{how} reviews are written and \emph{how fast} they are produced. A writing agent drafts each section sequentially, with limited ability to revise earlier passages once new evidence arrives later, so that errors in early structural decisions can propagate through the rest of the review. The same sequential bottleneck constrains the orchestration layer: when an agent must emit several search queries to fan out retrieval over distinct facets of the literature, autoregressive decoding produces them one token at a time, and downstream retrieval is dispatched only after generation completes. For an interactive research assistant, every one of these stages is on the critical path, and their latencies add up.

Diffusion language models \citep{li2022diffusionlm,austin2021structured,lou2024discrete} offer a complementary generation paradigm: text is refined through iterative denoising that updates an entire sequence in parallel across a small number of steps. Recent systems such as LLaDA \citep{nie2025llada}, Dream \citep{ye2025dream}, and Mercury \citep{inception2025mercury} suggest that diffusion LLMs can approach the quality of autoregressive models while offering advantages in generation speed and global coherence. Yet diffusion LLMs have largely been evaluated in standalone settings such as chat and coding; their integration into agentic pipelines that interleave planning, retrieval, and iterative refinement remains underexplored, and the practical latency consequences of that integration have not been measured directly.

We introduce \textbf{DiffResearch},\footnote{Previous presentation: ICML 2026 AI4Science (\url{https://openreview.net/forum?id=EGb0wlKWVz}). This version adds a controlled AR-backbone comparison, dFast analysis, language-specific diagnostics, and a source-grounding audit. Code: \url{https://github.com/alanrbtx/DiffResearch}.} an open deep research framework in which every agent is an instance of a diffusion LLM. Our aim is to measure what changes when the backbone is diffusion rather than autoregressive, with the rest of the system held fixed. The framework combines a standard agentic scaffold (intent classification, query reformulation, planning, retrieval, writing, and an optional judge) with a diffusion-based writing stage that drafts and refines the full review in parallel rather than section by section. It supports two operating modes: a \emph{full} mode that runs the base pipeline once, and an iterative subquery-decomposition (\emph{fast}) mode in which a diffusion decomposition agent produces the full subquery set within a single denoising pass, with usable subqueries read from intermediate denoising states before generation completes. Because all subqueries become available at once, their retrieval calls could in principle be dispatched together; in our controlled runs retrieval is rate-limited to one request per second, so we report the decomposition-stage speedup in isolation and leave parallel retrieval dispatch to future work. The framework exposes one agent abstraction with interchangeable backbones, so we can keep the whole scaffold fixed and change only the generator, and any difference can be attributed to the decoding paradigm rather than to unrelated system changes.

Our contributions are as follows:
\begin{itemize}
  \item \textbf{An open diffusion-native deep research framework} with two operating modes and a single agent abstraction whose backbone can be set to a diffusion or an autoregressive model, enabling controlled comparison.
  \item \textbf{A direct latency comparison} between diffusion and autoregressive backbones in an otherwise identical pipeline on Deep Research Bench, showing roughly $6\times$ end-to-end speedup concentrated in the planning and writing stages.
  \item \textbf{A fast decomposition mode} that extracts the subquery set from intermediate denoising states, cutting decomposition latency by about half ($1.99\times$ at the median over $100$ queries) while keeping downstream report quality; in our runs it also gives the best RACE overall score, but no additional end-to-end latency reduction over standard diffusion generation.
  \item \textbf{A system-level leaderboard snapshot} in which an instance of the same framework built on a stronger diffusion model ranks above several publicly listed commercial deep research systems, without isolating the effect of diffusion decoding.
\end{itemize}

\section{Related Work}

\paragraph{Deep research agents.}
Commercial systems \citep{openai2025deepresearch,google2024gemini,perplexity2025deepresearch} and open-source frameworks \citep{langchain2025odr,zheng2025deepresearcher} integrate multi-step reasoning, retrieval, and synthesis to produce long-form, citation-grounded outputs. Benchmarks such as Deep Research Bench \citep{li2025deepresearchbench}, DeepResearchGym \citep{du2025deepresearchgym}, and ResearchQA \citep{zheng2025researchqa} evaluate these systems on complex, multi-domain tasks. Most rely on autoregressive LLMs \citep{brown2020language,touvron2023llama,openai2023gpt4} and inherit the constraints of sequential decoding. Prior work has explored agentic scaffolds that decompose tasks into modular planning, retrieval, and evaluation components \citep{wang2024survey,xi2023rise}, judge models that assess output quality \citep{zheng2023judging}, and retrieval-augmented generation to ground outputs in external sources \citep{lewis2020retrieval,gao2023rag}.

\paragraph{Diffusion language models.}
Diffusion-based text models \citep{li2022diffusionlm,austin2021structured,lou2024discrete} generate through iterative denoising, refining an entire sequence in parallel across a few steps, and naturally support structure-first generation and iterative error correction \citep{li2022diffusionlm,gong2023diffuseq}. Recent open models such as LLaDA \citep{nie2025llada} and Dream \citep{ye2025dream}, and the Mercury family \citep{inception2025mercury}, show that diffusion LLMs can approach autoregressive quality while offering speed and global-coherence advantages. To our knowledge, no prior work has measured how diffusion decoding affects the end-to-end latency and report quality of an agentic deep research pipeline under a controlled backbone swap.

\paragraph{Efficient generation.}
A large literature improves generation efficiency through speculative decoding, caching, distillation, and parallel decoding. Our fast decomposition mode is complementary and is \emph{not} speculative decoding: it does not use a draft model, but instead reads the diffusion model's own intermediate denoising states and stops once a valid structured output is available.

\section{System Design}

\subsection{Multi-Agent Diffusion Pipeline}

DiffResearch is a multi-agent framework in which every agent is an instance of a diffusion LLM. The system is organized as a small, explicit pipeline rather than a monolithic chain; we deliberately avoid heavy orchestration abstractions \citep{chase2022langchain} that obscure data flow and make latency budgets and retry behavior hard to reason about. Each agent is a Python class that inherits from a common backbone and implements a single \texttt{generate} method. The backbone is interchangeable: a Hugging Face diffusion runtime serving \texttt{google/diffusiongemma-26B-A4B-it} for diffusion-native deployment, or a Hugging Face autoregressive runtime serving \texttt{google/gemma-4-26B-A4B-it} for the baseline. The controlled comparison in Section~\ref{sec:results} relies on this: changing the backbone leaves every prompt, retrieval call, and evaluation step unchanged.

Nine agents cover query formatting, planning, plan-checking, relevance filtering, information extraction, summarization, complexity assessment, query decomposition, and judging. Outputs are kept minimal and machine-parseable: binary flags for routing decisions, newline-separated query lists from the decomposition and judge agents, and free-form text only where downstream consumption requires it. Per-agent decoding budgets are exposed through the constructor; for example, the relevance and complexity agents use a short budget appropriate to their binary outputs, while the summarization agent uses a substantially larger budget to accommodate full review drafts.

The base pipeline has five core stages (intent classification, query reformulation, planning, retrieval, and writing), with judging as an optional sixth stage. The intent agent routes a query to a web channel or an academic channel and emits a label with a logged-but-not-propagated rationale. The reformulation agent rewrites the query into retrieval-friendly search strings \citep{wang2023query2doc,jagerman2023query} while preserving the original information need verbatim. The planning agent \citep{wei2022cot,yao2023react} produces a structured outline: main sections, two-to-four sub-points each, and an explicit list of key concepts to cover. The outline serves both as the writing target and as the reference for plan-checking. The writing agent drafts the review conditioned on the plan and the retrieved corpus through diffusion-style refinement. The judge \citep{zheng2023judging} compares the review against the plan and corpus along plan adherence, citation faithfulness \citep{gao2023citations,liu2023verifiability}, and coverage completeness.

\subsection{Retrieval Backends}

Retrieval grounds every claim and is treated as a first-class component. The web channel issues queries through the Serper API \citep{serper2024} and extracts page content into plain text. The academic channel issues parallel calls to the Semantic Scholar API \citep{kinney2023semanticscholar}, for broad peer-reviewed coverage and citation-graph signals, and the arXiv API \citep{arxiv2024api}, for timely preprints; results are deduplicated by DOI and arXiv identifier and merged into a single ranked list. In all cases retrieval is performed by deterministic API calls with no LLM in the loop, which keeps it fast, cacheable, and easy to ablate. For the controlled benchmark runs in this paper, retrieval is restricted to the Semantic Scholar channel with \texttt{results\_per\_query}$=2$, so these parallel academic calls are not exercised; retrieval is then a single rate-limited source, identical across backbones.

\subsection{Model Serving and Decomposition}

Both models are run through Hugging Face runtimes. The same pipeline code drives all three configurations (diffusion full, diffusion fast, autoregressive), and only the backbone and the decompose mode change. The decompose call accepts a \texttt{mode} of \texttt{full} or \texttt{fast}, and its output is accepted only when it parses as JSON with exactly the requested number of non-empty subqueries, which gives both modes the same correctness criterion.

\subsection{Operating Modes}

\paragraph{Full mode.}
Full mode is ordinary diffusion generation. The decomposition prompt is built and the model is run to completion; the final text is parsed as JSON and accepted if it contains exactly the requested number of subqueries. This is the correctness-oriented configuration.

\paragraph{Fast mode.}
Fast mode extracts the decomposition before final generation completes. The same prompt is issued, but a custom streamer inspects each intermediate denoising state: the draft is decoded, stripped of thinking-channel artifacts, scanned for a JSON object containing a \texttt{subqueries} list, and accepted as soon as a valid exact-count object appears. When early stopping is enabled, the streamer raises an exception that the decompose routine catches and returns the early subqueries instead of waiting for the final pass. This is \emph{not} speculative decoding: the model is unchanged, and the only difference is that we read its own intermediate denoised states. The tradeoff is that an early draft may contain duplicated fragments or rough wording, but for retrieval, topical coverage matters more than exact phrasing.

\section{Experimental Setup}

We evaluate on Deep Research Bench \citep{li2025deepresearchbench}, a benchmark of 100 PhD-level research tasks (50 Chinese, 50 English) spanning 22 fields, calibrated against a statistical analysis of over 96{,}000 real-world user queries. We follow the benchmark's official RACE evaluation framework, which scores reports along comprehensiveness, insight, instruction-following, and readability, and reports an aggregate overall score; here RACE is run with cleaning, using Gemini~2.5~Pro as the evaluator. RACE scores in our controlled runs are reported on a $0$--$1$ scale.

For the controlled comparison we use two backbones of comparable size in the same pipeline: the diffusion model \texttt{google/diffusiongemma-26B-A4B-it} (denoted \textbf{dFull} and \textbf{dFast} for its full and fast decomposition modes) and the autoregressive model \texttt{google/gemma-4-26B-A4B-it} (\textbf{AR}). Both are run through Hugging Face runtimes on an H200 pod, with retrieval performed locally. All three configurations share the same agent code, retrieval (Semantic Scholar only, \texttt{subquery\_count}$=4$, \texttt{results\_per\_query}$=2$, one request per second, per-request timeout), and RACE evaluation; only the generator backbone and the decompose mode differ. AR uses standard left-to-right decoding with no draft-state stream. We additionally report a separate isolated study of the decomposition stage on 100 synthetic-but-realistic research queries, instrumented directly through the diffusion streamer with model loading excluded from all timings.

We additionally analyze the stored run artifacts without regenerating reports, repeating retrieval, or calling the evaluator. For every successful report we measure total non-whitespace characters, numeric inline citations, and Markdown content headings, and split these diagnostics and saved RACE scores by language. We also conduct a configuration-blinded manual source-grounded audit of 10 reports from common-success tasks (5 English, 5 Chinese; dFull $n=3$, dFast $n=3$, AR $n=4$ after unblinding). Prompt requirements are fixed before reading reports; five material claims per report and all inline citation mentions are checked against the exact retrieved evidence available to the writer. This diagnostic is not expert human evaluation or an open-world factuality test.

\section{Results and Discussion}
\label{sec:results}

\subsection{End-to-End Latency}

Table~\ref{tab:e2e} reports end-to-end latency over all 100 tasks of Deep Research Bench. Both diffusion configurations are about $6\times$ faster than the autoregressive baseline on the mean (AR $71.45$\,s vs.\ dFull $11.89$\,s and dFast $11.97$\,s), and the gap is consistent across the median and tail (p90 $111.87$\,s for AR vs.\ $16.13$\,s and $16.33$\,s). dFast does not improve end-to-end latency over dFull in these runs: its observed mean and p90 are slightly higher. Its benefit is therefore confined to the decomposition stage, which is small relative to planning, retrieval, and writing and is isolated in Section~\ref{sec:decomp}.

\begin{table}[t]
  \centering
  \small
  \begin{tabular}{lrrrr}
    \toprule
    \textbf{Mode} & \textbf{mean} & \textbf{p50} & \textbf{p90} & \textbf{vs AR} \\
    \midrule
    AR    & 71.45 & 48.78 & 111.87 & $1.0\times$ \\
    dFull & 11.89 & 11.87 & 16.13 & $6.0\times$ \\
    dFast & 11.97 & 12.07 & 16.33 & $6.0\times$ \\
    \bottomrule
  \end{tabular}
  \caption{End-to-end latency (seconds) over the 100 Deep Research Bench tasks. Both diffusion modes are about $6\times$ faster than AR, while dFast provides no additional end-to-end reduction over dFull.}
  \label{tab:e2e}
\end{table}

\subsection{Where the Time Goes}

Table~\ref{tab:stages} breaks latency down by stage. The two model-bound stages dominate the AR cost, and they are where diffusion helps most. Planning drops from $39.32$\,s to about $3.6$--$3.7$\,s ($10.7\times$), and article writing drops from $59.78$\,s to about $5.1$--$5.3$\,s ($11.7\times$). Retrieval is essentially unchanged across backbones, as expected, since it is a shared deterministic Semantic Scholar call rather than a model call; it is slightly slower in the diffusion runs only because more tasks reach the retrieval stage. Decomposition is small in absolute terms, but it is the one stage where the two diffusion modes differ, which we examine next.

\begin{table}[t]
  \centering
  \small
  \begin{tabular}{lrrr}
    \toprule
    \textbf{Stage} & \textbf{AR} & \textbf{dFull} & \textbf{dFast} \\
    \midrule
    Planner      & 39.32 & 3.68 & 3.64 \\
    Decomposer   & 3.69  & 0.62 & 0.36 \\
    Retrieval$^\dagger$ & 3.94  & 4.83 & 4.59 \\
    Writer$^\ddagger$   & 59.78 & 5.11 & 5.29 \\
    \bottomrule
  \end{tabular}
  \caption{Mean per-stage latency (seconds). $^\dagger$Retrieval is a shared Semantic Scholar call, not a model call. $^\ddagger$Writer is measured over tasks that reached the writing stage. The model-bound planner and writer stages drive the $\sim$$10$--$12\times$ stage-level speedups.}
  \label{tab:stages}
\end{table}

\subsection{Accelerating Decomposition}
\label{sec:decomp}

Fast mode targets the decomposition stage specifically by reading intermediate denoising states. In the full pipeline this lowers mean decomposition latency from $0.62$\,s (dFull) to $0.36$\,s (dFast), about $1.7\times$, and against AR's $3.69$\,s it is roughly $10\times$. To characterize the mechanism in isolation, we ran a 100-query probe of the decomposition stage alone, with the model asked to return exactly six search subqueries as JSON, timing each query from the start of generation with model loading excluded. A valid exact-count object appeared during denoising for all 100 queries, with a mean of $8.99$ draft steps (median $9$, range $6$--$17$).

Table~\ref{tab:decomp} reports the latency of three extraction points. Taking the \emph{first} exact-count draft is $1.99\times$ faster at the median than waiting for full generation ($243.4$\,ms vs.\ $484.9$\,ms); requiring a \emph{stable} object (the same exact-count JSON observed for two consecutive states) is only $1.04\times$ faster at the median but reproduces the final answer exactly in 91 of 100 cases, versus 1 of 100 for the first draft.

\begin{table}[t]
  \centering
  \small
  \begin{tabular}{lrrrr}
    \toprule
    \textbf{Extraction point} & \textbf{Mean} & \textbf{P50} & \textbf{P90} & \textbf{P95} \\
    \midrule
    First exact-6    & 268.5 & 243.4 & 292.4 & 338.7 \\
    Stable exact-6   & 475.7 & 468.4 & 537.0 & 581.1 \\
    Full generation  & 499.7 & 484.9 & 579.8 & 629.9 \\
    \bottomrule
  \end{tabular}
  \caption{Isolated decomposition-latency probe over 100 queries (milliseconds; model loading excluded). Reading the first exact-count JSON from intermediate denoising states is $1.99\times$ faster at the median than full generation; the stable object is $1.04\times$ faster but matches the final answer in 91 of 100 cases.}
  \label{tab:decomp}
\end{table}

The takeaway is that diffusion draft states can be read as usable intermediate outputs for agent planning. For decomposition, exact final phrasing matters less than topical coverage, because the subqueries feed a retrieval call that tolerates wording noise, so the benchmark uses the first exact-count draft. The same reasoning does not hold for the final answer, where wording quality is the deliverable, which is why fast mode is scoped to decomposition and not to writing. In the current pipeline this is a stage-level optimization, not an additional end-to-end speedup.

\subsection{Report Quality}

A faster pipeline is only useful if quality holds. Table~\ref{tab:race} reports RACE scores under cleaning and the requested language-specific diagnostics. The diffusion fast mode obtains the best available-case overall score ($0.3969$), ahead of dFull ($0.3895$) by $0.0074$ and ahead of AR ($0.3904$) by $0.0065$. In the paired common-success analysis, every overall-RACE mean-difference interval includes zero in both languages. Replacing autoregressive decoding with diffusion, and extracting decomposition from draft states, does not lower report quality, and it delivers the latency reductions reported above.

\begin{table}[t]
  \centering
  \scriptsize
  \setlength{\tabcolsep}{3pt}
  \begin{tabular}{lrrrrr}
    \toprule
    \multicolumn{6}{l}{\emph{(a) RACE dimensions}} \\
    \textbf{Mode} & \textbf{Overall} & \textbf{Comp.} & \textbf{Insight} & \textbf{Inst.} & \textbf{Read.} \\
    \midrule
    dFull & 0.3895 & 0.3577 & 0.3518 & 0.4489 & 0.4328 \\
    dFast & \textbf{0.3969} & 0.3624 & 0.3556 & 0.4610 & 0.4412 \\
    AR    & 0.3904 & 0.3413 & 0.3612 & 0.4436 & 0.4438 \\
    \midrule
    \multicolumn{6}{l}{\emph{(b) Language diagnostics over successful reports (means)}} \\
    \multicolumn{2}{l}{\textbf{Mode/lang}} & \textbf{RACE} & \textbf{Chars} & \textbf{Cites} & \textbf{Heads} \\
    \midrule
    \multicolumn{2}{l}{dFull/en} & 0.3925 & 5014 & 6.67 & 10.40 \\
    \multicolumn{2}{l}{dFull/zh} & 0.3858 & 2147 & 3.25 & 10.96 \\
    \multicolumn{2}{l}{dFast/en} & 0.3909 & 4853 & 7.14 & 10.64 \\
    \multicolumn{2}{l}{dFast/zh} & 0.4016 & 2176 & 3.83 & 11.25 \\
    \multicolumn{2}{l}{AR/en}    & 0.3925 & 5847 & 10.05 & 10.75 \\
    \multicolumn{2}{l}{AR/zh}    & 0.3885 & 2410 & 8.52 & 12.52 \\
    \bottomrule
  \end{tabular}
  \caption{Quality diagnostics. Panel (a): RACE with cleaning and a Gemini~2.5~Pro evaluator. Panel (b): language-specific available-case means; Chars are complete-report non-whitespace characters, Cites are numeric inline citation groups, and Heads are H2--H6 content headings.}
  \label{tab:race}
\end{table}

\begin{samepage}
The descriptive length differences are much smaller than the latency gap: diffusion reports are about $10$--$17\%$ shorter than AR reports in the language-specific means, whereas the writer is about $11$--$12\times$ faster (Table~\ref{tab:stages}). Shorter outputs alone therefore cannot explain the writing-stage speedup.
\end{samepage}

\paragraph{Manual source-grounded audit.}
Across the 10 blinded reports, 14 of 50 sampled material claims (28\%) were fully supported by the retrieved evidence and 10 (20\%) were partially supported; 26 (52\%) were not established by the stored source packet. Of 76 inline citation mentions, 41 (53.9\%) fully entailed the adjacent claim, 21 (27.6\%) partially entailed it, and 14 (18.4\%) did not establish it. Eleven of 22 cited sources were strong or adequate for the specific use. Prompt-facet coverage was high overall (87.4\%; English 92.7\%, Chinese 82.2\%). The recurring failures were metadata-only or off-topic Semantic Scholar results, unsupported extrapolation, and missing quantitative evidence. These findings identify retrieval grounding as a shared system bottleneck rather than a decoding-specific effect; the per-mode audit samples are too small for comparative claims.

\subsection{System-Level Leaderboard Snapshot}

The controlled study above uses two open models of comparable size to isolate the effect of the backbone. Separately, to assess end-to-end system competitiveness rather than isolate a mechanism, we evaluated an instance of DiffResearch built on a stronger diffusion model (Mercury~\citep{inception2025mercury}) on the public Deep Research Bench leaderboard, where RACE is reported on a $0$--$100$ scale. As Table~\ref{tab:leaderboard} shows, this instance scores $48.03$ overall, ranking above several publicly listed commercial systems at the time of evaluation, including OpenAI Deep Research ($46.45$), Dr.\ Tulu ($45.49$), and Claude Research ($45.00$). Its strongest results are on insight ($48.20$) and readability ($47.40$), the two dimensions most directly tied to long-form synthesis. Because leaderboard entries differ in retrieval infrastructure, model scale, and engineering investment, this time-specific snapshot supports a system-level competitiveness claim only and does not attribute the ranking to diffusion decoding.

\begin{table}[t]
  \centering
  \small
  \begin{tabular}{lrrr}
    \toprule
    \textbf{System} & \textbf{Overall} & \textbf{Insight} & \textbf{Read.} \\
    \midrule
    DiffResearch (Mercury)        & \textbf{48.03} & 48.20 & 47.40 \\
    OpenAI Deep Research          & 46.45 & 43.73 & 47.22 \\
    UESTC-MBSE-RAAA               & 46.13 & 48.34 & 43.78 \\
    Dr.\ Tulu                     & 45.49 & 44.65 & 42.30 \\
    Claude Research               & 45.00 & 42.79 & 44.66 \\
    Kimi Researcher               & 44.64 & 41.97 & 45.59 \\
    Doubao Deep Research          & 44.34 & 40.56 & 44.69 \\
    LangChain Open DR             & 43.44 & 39.17 & 45.22 \\
    NVIDIA AIQ                    & 40.52 & 38.39 & 42.63 \\
    Tongyi DR 30B-A3B             & 40.46 & 34.44 & 44.27 \\
    \bottomrule
  \end{tabular}
  \caption{Public Deep Research Bench leaderboard snapshot (RACE, 0--100). The complete Mercury-backed DiffResearch system ranks above several listed commercial systems; this comparison does not isolate diffusion decoding.}
  \label{tab:leaderboard}
\end{table}

\subsection{Discussion}

Our results support a narrow but practical claim. With the backbone as the only change, diffusion decoding makes the model-bound stages of the pipeline, planning and long-form writing, about an order of magnitude faster, which gives a $6\times$ end-to-end speedup without lowering RACE quality. Diffusion also exposes intermediate denoising states that can be read mid-generation. For a structured intermediate output such as a decomposition, these drafts are usable well before the final pass, which adds a further $2\times$ on that stage in isolation but no additional end-to-end reduction in the current pipeline. The source-grounded audit shows that this decoding gain does not solve retrieval relevance or evidence sufficiency, which remain shared bottlenecks. Separately, the public leaderboard snapshot shows that the complete Mercury-backed DiffResearch system is competitive with listed commercial systems, but it does not attribute that ranking to diffusion decoding. Thus, the controlled study supports the decoding-paradigm latency claim, whereas the leaderboard supports end-to-end system competitiveness.

\section*{Limitations}

The end-to-end speedups are measured with the model served on an H200 pod and retrieval kept local. On hardware or model configurations where local diffusion inference becomes the bottleneck, or where retrieval cost is small relative to generation, the absolute latency gap will differ; we have not benchmarked under tight consumer-grade latency budgets. Fast mode is deliberately scoped to decomposition and does not reduce observed end-to-end latency relative to dFull in the current setup: the first exact-count draft rarely matches the final string exactly (1 of 100 in our probe), so it is appropriate only where downstream consumers tolerate wording noise, as retrieval does, and not for the final answer. The controlled comparison uses two open models of comparable size; while it holds the scaffold, prompts, retrieval, and evaluation fixed, it is a single backbone pair, and a broader sweep across model families and sizes would strengthen the attribution. Finally, the leaderboard comparison is against full systems rather than writer-isolated variants and reflects a snapshot in time; it should be read as a system-level result, not a causal attribution to diffusion decoding alone. Retrieval in these runs used Semantic Scholar only, which for some benchmark tasks did not return the sources the task needed; the academic channel can also use arXiv and the web channel, but both were disabled here. Parallel retrieval dispatch, which single-pass subquery generation enables in principle, is not evaluated here because retrieval is rate-limited to one request per second; quantifying it under a higher-throughput retrieval backend is left to future work.

\section*{Acknowledgments}

This paper was carried out as part of ITMO University project No.~666001 ``Laboratory of Multi-Agent Modeling and Adaptive Intelligence''.

\bibliography{custom}

\end{document}